\documentclass[10pt,twocolumn,letterpaper]{article}

\usepackage{cvpr}
\usepackage{times}
\usepackage{epsfig}
\usepackage{graphicx}
\usepackage{amsmath}
\usepackage{amssymb}

\usepackage{epsfig}
\usepackage{graphicx}
\usepackage{amsmath}
\usepackage{amssymb}
\usepackage{multirow}
\usepackage{color}
\usepackage{algorithmic}
\usepackage{algorithm}
\usepackage{booktabs}
\usepackage[table]{xcolor}
\usepackage{colortbl}
\usepackage{bm}
\usepackage{subcaption}
\usepackage{makecell}
\usepackage{pifont}

\usepackage{threeparttable}
\usepackage{stfloats}

\usepackage{enumitem}

\usepackage[hidelinks,breaklinks=true,bookmarks=false]{hyperref}

\cvprfinalcopy 

\def\cvprPaperID{****} 

\begin{document}


\title{Workshop on Autonomous Driving at CVPR 2021: \\Technical Report for Streaming Perception Challenge }

\author{Songyang Zhang$^{1,}$\thanks{equal contribution. This work is done when Songyang Zhang, Lin Song and Zheng Ge are research interns at Megvii Technology.}\quad Lin Song$^{2, *}$\quad Songtao Liu$^{4,*}$\quad Zheng Ge$^{3}$
\quad Zeming Li$^{4}$\quad Xuming He$^{1}$\quad Jian Sun$^{4}$\\

$^1$ShanghaiTech University\quad $^2$Xi'an Jiaotao University\quad $^3$Waseda University\quad $^4$Megvii Technology\\
{\tt\small sy.zhangbuaa@gmail.com, stevengrove@stu.xjtu.edu.cn,  jokerzz@fuji.waseda.jp}
\\
{\tt\small \{liusongtao,gezheng,lizeming,sunjian\}@megvii.com, hexm@shanghaitech.edu.cn}
}

\maketitle

\begin{abstract}
In this report, we introduce our real-time 2D object detection system for realistic autonomous driving scenario. Our detector is built on a new designed YOLO model, called YOLOX. On the Argoverse-HD dataset, our system achieves 41.0 streaming AP, which surpassed the second place by 7.8/6.1 on detection-only track/fully track, respectively. Moreover, equipped with TensorRT, our model achieves the 30FPS inference speed with a high-resolution input size (\textit{e.g.,} 1440$\times $2304). Code and models will be available at \href{https://github.com/Megvii-BaseDetection/YOLOX}{https://github.com/Megvii-BaseDetection/YOLOX}
\end{abstract}

\section{Overview}
Our goal is to build a fast and accurate 2D detector for autonomous driving scenario, we follow the YOLO series models and introduce several improvements to build our detection system. Below we first present the main idea of our method in Sec.~\ref{sec:structure}, followed by the network architecture used in the proposed detector in Sec~\ref{sec:structure}. Then, we describe the optimization strategy used to speedup the inference in Sec.~\ref{sec:inference}. Finally, we report the experiment and detailed ablation analysis in Sec.~\ref{sec:experiment}.

\section{Our Approach}\label{sec:method}
As the inference speed is quite important for this challenge, we adopt an internal new designed YOLO model, named YOLOX\cite{yolox2021}, for this task. 

Specifically, YOLOX\cite{yolox2021} follows the advanced data augmentation strategies of YOLOv4~\cite{bochkovskiy2020yolov4} and YOLOv5~\cite{yolov5}, such as mosiac, mixup. Then, we replace the anchor-based YOLOv5 detection head with an anchor-free  head and adopt a simplified version of OTA \cite{ota}, an advanced label assignment strategy, for training.

The above new designs simplify the current YOLO detectors and diminish many hyperparameters such as anchor shape, layer loss weight, \emph{etc}. Indeed, YOLOX achieves about 1.5\% higher AP than current YOLOv5 on COCO at the same speed. 

\section{Model Structure}\label{sec:structure}
We adopt the same C3 backbone as YOLOv5-L-P6. We refer the reader to YOLOX\cite{yolox2021} for the details of the overall model structure.

\section{Inference Optimization}\label{sec:inference}
To speedup the inference process for deployment, we adopt the TensorRT to generate the final model. Specifically, we convert the plain model trained with aforementioned strategy with \textit{Torch2TRT}~\cite{nvidiatorch2trt}. We fuse the image pre-processing operation and the prediction post-processing operation (NMS) into model forward function, which enables the whole inference process can be finished with a single function interface. 

Given the all-in-one inference interface, We simply the inference logic by modifying the provided toolkit, and  submit the generated predictions on validation set and 
test set to evaluation server .

\section{Experiments and Analysis}\label{sec:experiment}

\subsection{Training Configuration}
We implement the proposed YOLO X model with PyTorch~\cite{NEURIPS2019_9015}, and training the model on our internal deep learning computation platform. The overall learning consists of two stages: \textit{pre-training stage} and \textit{multi-dataset fine-tuning stage}.

\paragraph{Pre-training Stage.} We learn the YOLOX model by leveraging COCO 2017 dataset~\cite{lin2014microsoft}, which containes 135K images for traning and 5K images for valiation. We use the SGD with the momentum of 0.9. We use the cosine learning rate scheduler of 300 epochs with 5 epoch warmup. When pre-train COCO, we use 16 V100 GPUs (8 images/GPU). Therefor 128 images are utilized per mini-batch.

\paragraph{Multi-dataset Fine-tuning Stage.} Due to the limited data provided in Argoverse-HD~\cite{li2020towards} train set, we utilize several additional datasets to enhance the model capacity. 

Specifically, though there are 39384 images in train set, which mainly comes from only 65 video sequences. In other word, most of the images share the similar scene content, which makes the model easy to over-fitting. To tackle this issue and improve detector, we introduce BDD100K~\cite{yu2020bdd100k}, Cityscapes~\cite{cordts2016cityscapes} and nuScenes~\cite{caesar2020nuscenes} for training, which are all collected from the autonoumous driving scenario and have the similar scene content/categories with Argoverse-HD. We report the detailed statistics of the extra dataset in Tab-\ref{tab:dataset}.

We use the pre-trained model in stage-1 to initilize the detector and train the model with Argoverse-HD, BDD100K, Cityscape and nuScenes together in this stage. For the multi-dataset fine-tuning, we train the model with a learning rate at 0.01, 5 epoch for warmup and 20 epoch in total. We use 16 V100 GPU(6 images/GPU) with 96 images for a mini-batch.

\begin{table*}[t!]
	\centering
	\resizebox{0.8\textwidth}{!}{
		\begin{tabular}{l|c|c|c|c|c|c}
			\toprule
			    Dataset & Type &  Num Img(Train) & Num Img(Val) & Resolution & Num Class & Class Overlap\\
				\midrule
				COCO~\cite{lin2014microsoft} & Common & 118287 & 5000 & multi-scale & 80 & True\\ 
				\midrule
				Argoverse-HD~\cite{li2020towards} & Road &	39384  & 15062 & 1200$\times$1920 & 8 & True \\
				\midrule
				BDD100-K~\cite{yu2020bdd100k} & Road                    &  69863          & 10000  &  720$\times$1280 & 10 & True\\
				Cityscape~\cite{cordts2016cityscapes} & Road &                       2975          & 500  &  1024$\times$2048 & 19 & True \\
				nuScenes~\cite{caesar2020nuscenes} & Road & 67279 & 16445 & 900$\times$1600 & 25 & True\\ 
			\bottomrule
		\end{tabular}
}
	\caption{Statistics of the datasets used in the fine-tuning stage.}
	\label{tab:dataset}
\end{table*}

\begin{table*}[t!]
	\centering
	\resizebox{0.8\textwidth}{!}{
		\begin{tabular}{l|c|c|c|c|c|c}
			\toprule
			    Method & AP & AP$_{50}$ & AP$_{75}$ & AP$_{S}$ & AP$_M$ & AP$_L$  \\
				\midrule
				Strong Baseline (Single Argoverse-HD Dataset) & 35.4 & 59.9 & 34.2 & 18.3 & 34.2 & 48.4    \\
				+ COCO Pre-train  & 42.8 & 62.9 & 44.7 & 22.0 & 43.0 & 64.7\\
				+ Multi-dataset & 48.7 & 72.0 & 50.1 & 25.5 & 54.1 & 71.3    \\
				+ Large Scale Inference & 50.6 & 74.4 & 52.3 & 27.5 & 57.0 & 70.8 \\
				\midrule
				Online(on Val) & 40.2 & 68.9 & 39.4 & 21.5 & 42.9 & 53.9 \\
				Online(on Test) &  40.1 & 68.3 & 40.6 & 15.9 & 40.6 & 44.6\\
				Online$^\dagger$(on Test) & 41.0 & 69.5 & 41.2 & 15.1 & 41.8 & 47.7 \\
			\bottomrule
		\end{tabular}
}
	\caption{Detection road map of streaming perception challenge. We report the offline AP in the first block and the streaming AP in the second block. $\dagger$ means using validation set for fine-tuning. We use the 1440$\times$ 2304 size for large scale inference.}
	\label{tab:experiment}
\end{table*}
\subsection{Detection Road Map}
We report the detection performance in Tab.~\ref{tab:experiment}. Using COCO dataset for pre-training significantly improves our strong baseline with a gain of $7.4$ AP. Training with multiple datasets jointly further achieves $48.7$ AP. Moreover, evaluation with large-scale(1440$\times$ 2304) improves the performance to $50.6$. We achieve 40.2 on validation set in terms of the online evaluation metric equipped with TensorRT. By adding the validation set for fine-tuning the model, we achieve 41.0 AP(online) on test set.

\subsection{Latency vs Accuracy}
We have also reported the latency and the performance in terms of different image sizes used for inference, in Tab.~\ref{tab:ablation}
There exists a trade-off between the performance and latency, which is depended on the specific requirements of different task scenarios.

\begin{table}[t!]
	\centering
	\resizebox{0.32\textwidth}{!}{
		\begin{tabular}{l|c|c|c}
			\toprule
			    Width & Height & Speed(ms)  & AP   \\
				\midrule
				1440  &  2304 & 28.1 & 50.6 \\
				1280  &  2048 & 21.4 & 49.9 \\
				1200  &  1920 & 20.5 & 49.7 \\
				1120  &  1792 & 19.7 & 48.7 \\
				960   &  1536 & 16.0 & 46.3 \\
			\bottomrule
		\end{tabular}
}
	\caption{Performance v.s. Inference Latency}
	\label{tab:ablation}
\end{table}

{\small
\bibliographystyle{ieee_fullname}
\bibliography{egbib}
}

\end{document}